\pdfoutput=1

\documentclass[11pt]{article}

\usepackage{acl}

\usepackage{times}
\usepackage{latexsym}

\usepackage[T1]{fontenc}

\usepackage[utf8]{inputenc}

\usepackage{microtype}

\usepackage{inconsolata}

\usepackage{dblfloatfix}
\usepackage{graphicx}
\usepackage{placeins}
\usepackage{booktabs}
\usepackage{arydshln}
\usepackage{nicematrix}

\usepackage{hyphenat}
\usepackage{xspace}

\newcommand{\newxspacecommand}[2]{\newcommand{#1}{#2\xspace}}

\newxspacecommand{\sapbert}{SapBERT}
\newxspacecommand{\coder}{CODER}
\newxspacecommand{\twimedPM}{\twimed-PM}
\newxspacecommand{\twimedTW}{\twimed-TW}

\newxspacecommand{\psytar}{PsyTAR}
\newxspacecommand{\cadec}{CADEC}
\newxspacecommand{\twimed}{TwiMed}
\newxspacecommand{\smmfh}{SMM4H}

\newxspacecommand{\biolord}{BioLORD}
\newxspacecommand{\biolordPMB}{\biolord-PMB}
\newxspacecommand{\biolordSTAMB}{\biolord-STAMB2}
\newxspacecommand{\biolordSTAMBsts}{\biolordSTAMB-STS}
\newxspacecommand{\biolordSTAMBstsII}{\biolordSTAMB-STS2}

\newxspacecommand{\pretraining}{pre-training}
\newxspacecommand{\pretrained}{pre-trained}
\newxspacecommand{\finetuning}{fine-tuning}
\newxspacecommand{\finetuned}{fine-tuned}

\title{Boosting Adverse Drug Event Normalization on Social Media: General-Purpose Model Initialization and Biomedical Semantic Text Similarity Benefit Zero-Shot Linking in Informal Contexts} 

\author{François REMY \\
  University of Ghent \\
  \texttt{francois.remy} \\
  \texttt{@ugent.be} \\\And
  Simone Scaboro \\
  University of Udine \\
  \texttt{scaboro.simone} \\
  \texttt{@spes.uniud.it} \\\And
  Beatrice Portelli \\
  University of Udine \\
  \texttt{portelli.beatrice} \\
  \texttt{@spes.uniud.it} \\}

\begin{document}
\maketitle

\begin{abstract}
Biomedical entity linking, also known as biomedical concept normalization, has recently witnessed the rise to prominence of zero-shot contrastive models. However, the \pretraining material used for these models has, until now, largely consisted of specialist biomedical content such as MIMIC-III clinical notes \citep{johnson-2016-MIMIC} and PubMed papers \citep{sayers-2021-cz,gao-2020-pile}. While the resulting in-domain models have shown promising results for many biomedical tasks, adverse drug event normalization on social media texts has so far remained challenging for them \citep{portelli-etal-2022-generalizing}. In this paper, we propose a new approach for adverse drug event normalization on social media relying on general-purpose model initialization via \biolord \citep{remy-etal-2022-biolord} and a semantic-text-similarity \finetuning named STS.
Our experimental results on several social media datasets demonstrate the effectiveness of our proposed approach, by achieving state-of-the-art performance. Based on its strong performance across all the tested datasets, we believe this work could emerge as a turning point for the task of adverse drug event normalization on social media and has the potential to serve as a benchmark for future research in the field. 
\end{abstract}

\section{Introduction}

\begin{figure}[t]
\centering
\includegraphics[width=\linewidth]{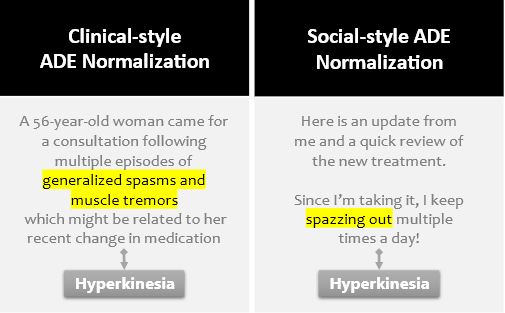}
\caption{\vspace{-0.05cm}Normalization of concepts in the clinical domain made large progresses, but social media content remains more challenging due to informal language.
\vspace{-0.25cm}}
\label{fig:norm_comparison}
\end{figure}

Adverse drug events (ADEs) are unexpected and possibly undocumented negative effects related to the correct use of a drug, and they have the potential to result in serious harm to patients. 
ADEs can also increase hospitalization costs, 
reduce patient satisfaction, and erode trust in the health care system.
For these reasons, ADEs are a major concern for patients, healthcare providers, and regulators. 

However, detecting and reporting emerging ADEs (a process known as pharmacovigilance) is not an easy task \citep{pappa-2019-social-media}. Most of the information about ADEs is buried in unstructured text sources, such as medical case reports, social media posts, or online reviews \citep{audeh-2020-vigi4med}.
The two latter sources often contain informal language, abbreviations, slang, or misspellings, that make machine learning models unable to accurately extract and normalize ADEs present within them, a process known as biomedical concept normalization. Models trained exclusively on clinical data are particularly likely to be affected (see Figure \ref{fig:norm_comparison}).
This is a real concern, as mapping ADEs to standardized ontologies, such as MedDRA \citep{meddra} or SNOMED CT \citep{snomedct}, is an important step to facilitate the analysis and comparison of ADE data across different sources and domains \citep{adel-2019-interoperability}.

In a short time span, between the years 2020 and 2022, the field of biomedical concept normalization has seen significant advancements with the introduction of self-supervised contrastive models. Originally introduced by \citet{chen-2020-simple} for computer vision, these models are trained to produce identical latent representations for multiple views of a same concept, yet contrasted representations for each concept. In the biomedical domain, these views are usually constructed based on biomedical ontologies, by pairing canonical names of a concept with some of their known synonyms. 

State of the art biomedical entity normalization is now dominated by models relying on this technique such as BioSyn \citep{sung-etal-2020-biomedical}, \coder \citep{yuan-2022-coder}, and \sapbert \citep{liu-etal-2021-self}. What makes these models extremely versatile is that it is possible to encode a new set of target concepts at inference time, which means that using the same model is possible irrespective of the target ontology, enabling smooth system updates.

\section{Our contributions}
\subsection{General-Purpose Initialization}
All models cited thus far were initialized from language models pre-trained on text from the biomedical domain, as this is thought to improve entity normalization performance somewhat in the clinical domain. In this paper, we propose a new approach for adverse drug event normalization on social media by employing \biolord, a general-purpose model initialization approach pioneered by \citet{remy-etal-2022-biolord}. We hypothesize that its pre-training on general texts will help tremendously in understanding the informal language used on social media, while previous state of the art models struggled at that specific task, due to the domain shift between clinical and social media languages. 

\subsection{MedSTS Finetuning}
In addition, after noticing that semantic-text-similarity finetuning helps achieving better performance, we improve this new approach even further by incorporating two distinct semantic-text-similarity (STS) \finetuning phases to the training, both before and after the BioLORD \pretraining. We release the improved BioLORD-STS model as part of this paper, and show that it achieves a performance far exceeding the current state-of-the-art.

\section{Methodology}
\label{sec:method}

\begin{figure}[t]
\centering
\includegraphics[width=\linewidth]{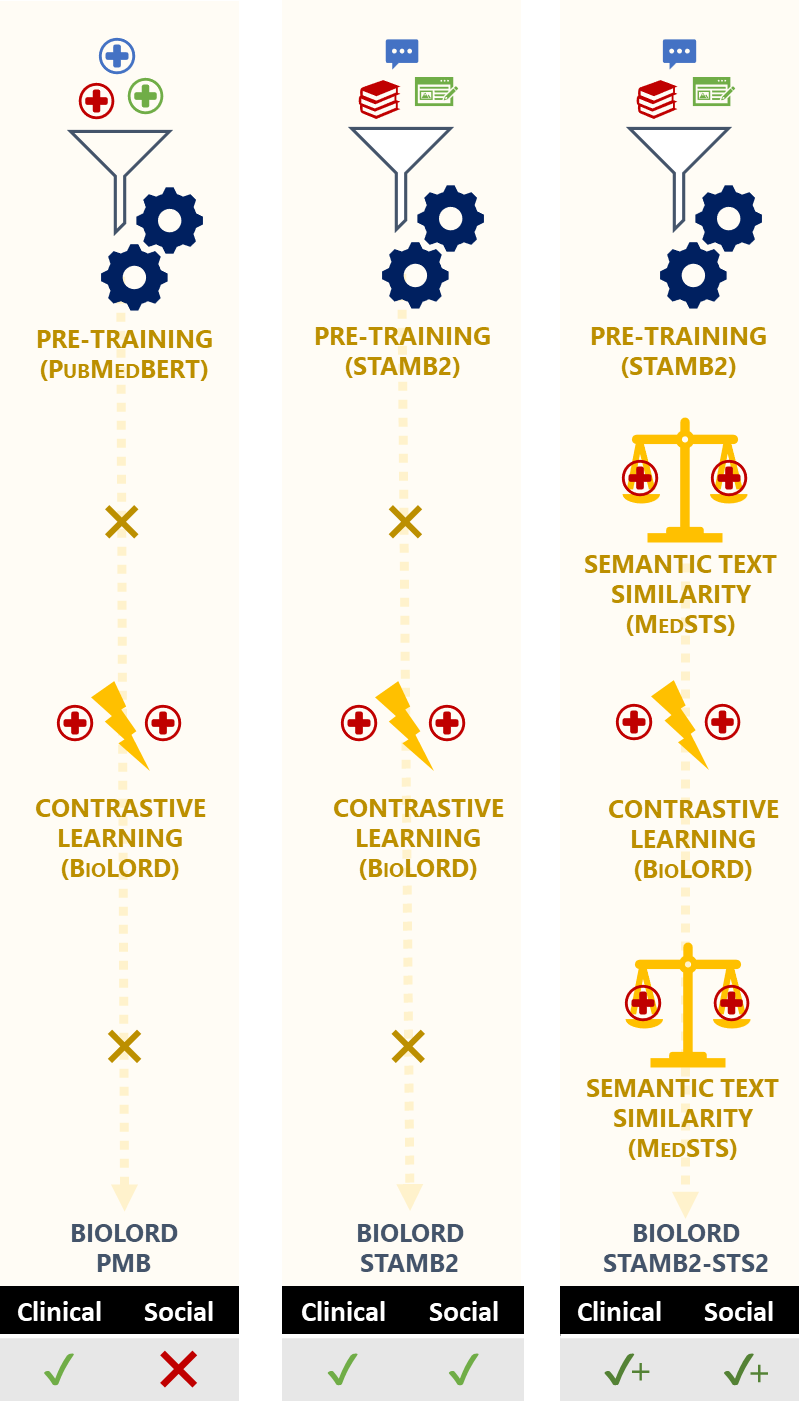}
\caption{Schema of the \pretraining and \finetuning steps of the four candidate models: \biolordPMB, \biolordSTAMB, and \biolordSTAMBstsII.
}
\label{fig:architectures}
\end{figure}

In this paper, we set out to show that general-purpose models \finetuned on biomedical definitions perform better, for the task of ADE normalization in the social media, than state-of-the-art models trained exclusively on biomedical corpora. 

Our hypothesis is based on the following observations derived from the extensive ablation studies performed in the \biolord \pretraining paper \citep{remy-etal-2022-biolord}: applying the \biolord \pretraining strategy on a general-purpose model can help the model learn to generalize across different writing styles, including non-clinical ones that are rare in biomedical corpora; meanwhile, possessing a strong biomedical knowledge at initialization time did not appear essential to achieve good performance when using the \biolord \pretraining.

In a final experiment, we also verify that biomedical text similarity is a useful \pretraining step to apply before training \biolord-type models.

We benchmark all models on four social media datasets: \cadec \citep{karimi-2015-cadec}, \psytar \citep{zolnoori-2019-psytar}, \smmfh \citep{weissenbacher-etal-2019-overview}, and \twimed \citep{alvaro-2017-twimed}, which are further described in Section \ref{sec:datasets}.

Through this extensive benchmarking, we aim to demonstrate that our proposed approach significantly outperforms the previous state of the art 
in informal contexts, which cover a wide range of social media activities (such as: forum messages, online reviews, and tweets). 

\subsection{Candidate Models}

To disentangle the impact of the base model initialization from the impact of the \biolord \pretraining strategy, we consider two different base models: STAMB2\footnote{\url{https://huggingface.co/sentence-transformers/all-mpnet-base-v2}} \citep{reimers-2019-sentence-bert}, the same general-purpose model used in the \biolord paper, and PubMedBERT \cite{gu-etal-2020-pubmedbert}, a robust domain-specific model \pretrained on medical texts. The resulting models are named \textbf{\biolordSTAMB} and \textbf{\biolordPMB}.

We also analyze the effect of \finetuning the models on a medical semantic text similarity task (STS) in addition to the \biolord \pretraining. We do so by \finetuning some of the models on the MedSTS task \citep{wang-2020-medsts}, using the same hyperparameters described in the \biolord paper.
The base STAMB2 model is \finetuned for STS before applying the \biolord \pretraining.
This model then undergoes a second stage of STS \finetuning, resulting in \textbf{\biolordSTAMBstsII}.

Figure \ref{fig:architectures} illustrates the differences between the proposed models.

\subsection{Baseline Models}

We choose two BERT-based models trained with contrastive learning strategies as baselines: \textbf{\coder} \citep{yuan-2022-coder} and \textbf{\sapbert} \citep{liu-etal-2021-self}.
They are among the best dataset-agnostic models for medical term embeddings at the time of writing. Both of them are trained on the UMLS ontology \citep{umls} and were tested on several term normalization datasets, showing promising results. \sapbert was the first large-scale contrastive model to leverage UMLS and is based on PubMedBERT. It is trained by using UMLS synonyms to create contrastive pairs. \coder, on the other hand, leverages both term-term pairs and term-relation-term triples.

\subsection{Datasets}
\label{sec:datasets}

We evaluate all the candidate and baseline models using four medical entity normalization datasets containing ADEs. All of them contain informal texts coming from different social media platforms. One of the datasets also contains a subset of formal samples (\twimedPM). We include this subset in our experiments to verify that all the tested models perform well on ADE normalization in the clinical domain too.

The \textbf{\twimed} dataset \citep{alvaro-2017-twimed} provides a comparable corpus of texts from PubMed (abstracts) and Twitter (posts), allowing researchers in the area of pharmacovigilance to better understand the similarities and differences between the language used to describe disease and drug-related symptoms on PubMed (\textbf{\twimedPM}, clinical domain) and Twitter (\textbf{\twimedTW}, social media domain). Both sets of data contain 1000 samples.

The {CSIRO Adverse Drug Event Corpus} (\textbf{\cadec}) dataset \citep{karimi-2015-cadec} is a corpus of user-generated reviews of drugs that has been annotated with adverse drug events (ADEs) and their normalization. It contains 1250 posts from a medical forum, which were annotated by a team of experts from the University of Arizona.

The {Psychiatric Treatment Adverse Reactions} (\textbf{\psytar}) dataset \citep{zolnoori-2019-psytar} contains patients’ expression of effectiveness and adverse drug events associated with psychiatric medications, originating from a sample of 891 drugs reviews posted by patients on an online healthcare forum. 

The {Social Media Mining for Health Applications} (\textbf{\smmfh}) dataset \citep{smm4h-2020-social} is a dataset for Adverse Drug Event (ADE) normalization. It was used in the \smmfh 2020 shared task on ADE normalization. The aim of the subtask was to recognize ADE mentions from tweets and normalize them to their preferred term in the MedDRA ontology. The dataset includes 1212 tweets containing ADEs.

For each evaluated dataset, we perform zero-shot entity normalization using a setup identical to \citet{portelli-etal-2022-generalizing} with four test splits.

\begin{table*}[!t]
\centering
\resizebox{\linewidth}{!}{
\begin{tabular}{lllllll}
\toprule
{} & {{\textbf{\twimedPM}}} & {{\textbf{\cadec}}} & {{\textbf{\psytar}}} & {{\textbf{\smmfh}}} & {{\textbf{\twimedTW}}} \\
\toprule
\small{\textsc{\textbf{\coder}}} & 65.31 ± 1.85 & 35.29 ± 1.27 & 52.40 ± 0.71 & 33.14 ± 1.28 & 42.80 ± 2.06 \\
\small{\textsc{\textbf{\sapbert}}} & 70.05 ± 1.56 & 40.42 ± 1.27 & 64.82 ± 1.36 & 43.37 ± 1.07 & 48.29 ± 2.85 \\
\midrule
\small{\textsc{\textbf{\biolordPMB}}} & 69.99 ± 1.87 & 58.23 ± 0.36 & 60.22 ± 0.84 & 41.80 ± 2.24 & 47.14 ± 2.21 \\
\small{\textsc{\textbf{\biolordSTAMB}}} & 70.44 ± 1.19 & 58.69 ± 0.97 & 64.70 ± 0.76 & 46.51 ± 2.08 & 48.46 ± 1.53 \\
\midrule
\small{\textsc{\textbf{\biolordSTAMBstsII}}} & \underline{70.60} ± 1.19 & \underline{60.28} ± 0.80 & \underline{65.49} ± 0.74 & \underline{47.33} ± 1.42 & \underline{50.57} ± 1.72 \\
\bottomrule
\end{tabular}
}
\caption{Accuracy@1 of the evaluated models on all the datasets. Datasets are ordered according to the formality of their language, from more formal (\twimedTW) to more informal (\smmfh and \twimedTW).}
\label{tab:results}
\end{table*}

\section{Results}

We report the zero-shot evaluation results of the various models in Table \ref{tab:results}.

Looking at the results on \twimedPM, the samples coming from the clinical domain, we observe that almost all of the models have a similar performance (between 69.99 and 70.60), showing that all models (general-purpose or in-domain) can reach a good performance on formal datasets.

The gap in performance of \coder and \sapbert between \twimedPM and all the social-media datasets highlights the existence of a significant difference in language distribution between texts in the clinical domain, and the less formal texts found in online reviews or social media.

If we focus on the models trained using only the \biolord \pretraining, we can see that they perform better than the two state-of-the-art baseline alternatives across all the datasets. In particular, \biolordSTAMB significantly outperforms \sapbert on \cadec (58.69 vs 40.42) and \smmfh (46.51 vs 43.37). We also observe that \biolordSTAMB, the general-domain model, outperforms \biolordPMB, the domain-specific variant, proving that the findings of the original \biolord paper extend to the social media domain.


Our results also highlight that the newly-introduced 
\biolordSTAMBstsII manages to move the needle even further (with an average accuracy gain of 1 point with respect to \biolordSTAMB), indicating that priming general-purpose models (STAMB2) for biomedical text understanding (STS) before and after the \biolord \pretraining enables to achieve better performance.

On the \cadec dataset in particular, our \biolord family of models achieves zero-shot accuracy@1 of above 60\% for Preferred Term classification. To the best of our knowledge, this is by far the best zero-shot performance ever reported for this dataset.

This seems to confirm that ADE normalization will continue to move towards self-supervised contrastive models, as these models perform well, are very versatile, and can be used to map concepts to any new updated ontology at test time without requiring any retraining. In a field where such models are expected to continue to thrive, the improvements proposed in this paper should be particularly of interest to other researchers.

\newpage
\section{Conclusion}

In this paper, we confirmed that \biolord is an effective \pretraining strategy for biomedical entity normalization. We were additionally able to show that applying \biolord on general-purpose models like STAMB2 provides additional benefits, and that these benefits are more important for sources originating from social media than from clinical notes. Finally, we report that STS-\finetuning of models both before and after undergoing the \biolord \pretraining can bring additional benefits even in the ADE normalization task, especially in the case when the source originates from social media documents.

\section*{Limitations}

This paper did not investigate the impact of the proposed \pretraining strategies on ADE identification, the task of finding ADE mentions in a text.

We also did not investigate the impact of \finetuning models on the task, although we have performed some preliminary experiments on this, which seem to confirm the conclusions for zero-shot models apply to \finetuned models as well.

\section*{Ethics Statement}

The authors do not foresee that their work would raised any particular ethical concern.

\clearpage
\bibliography{anthology,custom}
\bibliographystyle{acl_natbib}

\appendix

%

\end{document}